\documentclass[conference]{IEEEtran}
\IEEEoverridecommandlockouts
\usepackage{cite}
\usepackage{amsmath,amssymb,amsfonts}
\usepackage{algorithmic}
\usepackage{graphicx}
\usepackage{algorithm}
\usepackage{subcaption}
\usepackage{multirow}
\usepackage{booktabs}
\usepackage{textcomp}
\usepackage{xcolor}
\usepackage{makecell}
\def\BibTeX{{\rm B\kern-.05em{\sc i\kern-.025em b}\kern-.08em
    T\kern-.1667em\lower.7ex\hbox{E}\kern-.125emX}}
\DeclareUnicodeCharacter{0301}{\'{e}}
\begin{document}

\title{Personalized Federated Learning via Gradient Modulation for Heterogeneous Text Summarization \thanks{*Corresponding author: Jianzong Wang, jzwang@188.com.}
}

\author{\IEEEauthorblockN{Rongfeng Pan\IEEEauthorrefmark{0}, Jianzong Wang\IEEEauthorrefmark{0*}, Lingwei Kong\IEEEauthorrefmark{0}, Zhangcheng Huang\IEEEauthorrefmark{0}, and Jing Xiao\IEEEauthorrefmark{0}}
\IEEEauthorblockA{\textit{Ping An Technology (Shenzhen) Co., Ltd., Shenzhen, China}
}}

\maketitle

\begin{abstract}
Text summarization is essential for information aggregation and demands large amounts of training data. However, concerns about data privacy and security limit data collection and model training. To eliminate this concern, we propose a federated learning text summarization scheme, which allows users to share the global model in a cooperative learning manner without sharing raw data. 
Personalized federated learning (PFL) balances personalization and generalization in the process of optimizing the global model, to guide the training of local models. However, multiple local data have different distributions of semantics and context, which may cause the local model to learn deviated semantic and context information. In this paper, we propose FedSUMM, a dynamic gradient adapter to provide more appropriate local parameters for local model.  Simultaneously, FedSUMM uses differential privacy to prevent parameter leakage during distributed training. Experimental evidence verifies FedSUMM can achieve faster model convergence on PFL algorithm for task-specific text summarization, and the method achieves superior performance for different optimization metrics for text summarization.
\end{abstract}

\begin{IEEEkeywords}
Text Summarization, Personalized Federated Learning, Privacy Protection, Privacy-Preserving Computation
\end{IEEEkeywords}

\section{Introduction}
Text summarization is an important research area of information aggregation and natural language processing, which generates a condensed and concise summary from a long text \cite{cao2022hallucinated}\cite{nguyen2021enriching}, Real-world text summarization services rely on user data located in personal devices and large data silos between organizations. It is often limited by the high labor cost and time-consuming data construction. Meanwhile, with the release of data regulations, the issue of data privacy protection has received widespread attention. Data are often discretely distributed among individual users and organizations as isolated data islands\cite{ding2022federated}. Therefore, it is necessary to research and achieve a text summarization model with better performance and ensures user data privacy.

Federated learning (FL) has become a machine learning paradigm\cite{wang2021secure}, which can train models on various decentralized devices or servers, and it is a solution for constructing a joint-trained model while preserving data privacy and participants can keep data locally without exchanging \cite{FedEWA}. Federated learning was proposed for the first time by McMahan et al. \cite{mcmahan2017communication}, the learning task settled by loose jointing of participating devices (called clients) coordinated by a central server. 

Limited by the problem of non-interoperability of data between different clients, it is challenging for federated learning \cite{SunIJCNN} to achieve a balanced and optimized global model with non-i.i.d clients. Polato et al. \cite{PolatoIJCNN} pointed out the direction of degradation for all clients without sacrificing performance by attacking the expansion loss. Each client is personalized to find a robust and fair optimization method \cite{marfoq2021federated}. However, these methods aggregate a global model for different clients, fragile generalization performance when the non-II.D. Data exists textual semantic deviation.

Traditional FL algorithms suffer from the statistical heterogeneity of local non-II.D. data \cite{tang2022personalized}\cite{XuIJCNN}. However, personalized federated learning (PFL) seeks to address the problem by training a globally optimal shared model that can balance various local models. Many PFL methods are used to address the challenge from heterogeneous datasets, involving random sampling consensus \cite{li2022federated} is used to consider the imbalance between different clients (including self-supervised, unsupervised, and semi-supervised). The FedAMP algorithm with federated attentive message passing is proposed \cite{huang2021personalized} to coordinate the cross-island problem with similar data. Although PFL outperforms traditional FL, these methods are only indiscriminate optimization of parameters through a global model.

The intuition is that: distributional differences in semantics and contexts in local data. Therefore, there is a potential bottleneck in current PFL methods. When the global model transmits a single generalization loss and gradient parameters, it is easier to optimize the irrelevant text semantics in the local model due to differences in the distribution of local data, resulting in the decreased overall performance of the global model. Our work proposes that a personalized federated learning framework with a dynamic gradient adapter is used to deal with non-optimal optimization of global and local models due to text heterogeneity in text summarization tasks.

Our main contributions are as follows: 
\begin{itemize}
    \item Considering data scarcity and privacy protection, we investigate text summarization under federated learning, which decouples the need to aggregate data for joint training with multiple parties.
    \item We found that the semantic distribution deviation caused by text heterogeneity is often prone to unbalanced optimization of federated learning aggregation algorithms.
    \item We proposed the PFL algorithm with a dynamic gradient adapter: FedSUMM, which addresses the optimization imbalance problem and seek a better balance between personality and generalization during training.
\end{itemize}

\section{Related Work}
Considering that the Chinese text summarization tasks can hardly achieve better generalization performance due to the scarcity of data. At the same time, data privacy and security lead cannot be expanded by collecting data. Some recent works \cite{GaoSummarizing}\cite{UtamaFalsesum}\cite{LoemExtraPhrase} try to generate more training samples to alleviate this data scarcity problem. Back-translation is favored in construct pseudo-training predictions for augmentation purposes \cite{MauryaZmBART}. In recent years, Junxian et al. \cite{Junxianiclr} put forward a self-training approach to make the model generate target sentences from the source corpus. Magooda et al. \cite{MagoodaMitigating} believes that the data constructed by these methods often have similar data distribution characteristics to the original data.

Federated Learning was originally introduced in the seminal work of McMahan et al. In their work, the training task settled by loose jointing of participating equipment (called clients) coordinated by a central server. Many federated training approaches have been proposed in NLP. such as mobile keyboards prediction \cite{Hardkeyboard}, heterogeneous text classification\cite{Lizhengyang}, and mixed-domain translation models\cite{Peymantranslation}. However, only some jobs provide a federated solution tailored for text summarization tasks.

Recent works in federated learning applied to NLP can fall into two major aspects: On the one hand, data privacy protection \cite{WellerPretrained} primarily to overcome the problems caused by the data isolation limitations. On the other hand, it addresses data heterogeneity\cite{TanFedproto}. Liang et al. \cite{LiangModular} proposed to extract knowledge from a set of heterogeneous devices and non-IID data generated from its users. It was considered that the performance of aggregation algorithms for federated learning is due to the semantic distribution deviation caused by text heterogeneity. Personalized federated learning \cite{ChenPFL} has been widely studied to handle heterogeneity, which focuses on training a globally optimal shared model that can balance various local models.

Unlike previous studies, in this paper, we propose injecting a personalized federated setting for heterogeneous text summarization tasks and providing fine-grained generalization for personalized text summarization models.

\section{Proposed method}
In this section, we introduce text summarization models Bart and CopyTransformer use federated learning for distributed training and propose personalized federated learning with dynamic gradient adapter FedSUMM, which can adaptively modulate the gradient.

\subsection{Federated Bart}
To protect the security and privacy of the original data of the $n$ parties, we combine the traditional FL algorithm FedAvg into Bart for distributed training. Bart's pre-trained model combines bidirectional and autoregressive Transformer, which have achieved remarkable results in a lot of NLP applications. It was introduced in the paper\cite{lewis2020bart} by Lewis et al.

As shown in Fig.1, our federated summary model divides into two parts from top to bottom: The first part is a central server whose purpose of realizing the coordinated training of parameters among clients without sharing the private data of all clients. Its primary function is to receive parameters $w^{}_{n}$ obtained by clients training in each round of communication, and average these parameters through the FedAvg, finally return the updated parameter $w^{update}_{t}$ to local models for processing the next training session. The second part is local $n$ clients. Each client trains on its private data, where the input data is \textless private data, abstract\textgreater   pairs. In the experiment, we set n $\in$ [2,200]. The model structure and the number of layers of each partner are the same. In each round of communication, the local $n$ clients send the model parameters obtained in this iterative training to the server side for parameter averaging. Furthermore, with updated server-side parameters, the next round of iterative optimization is performed until the characteristic of the integrated overall model is enhanced.

\begin{figure}[htb]
\begin{minipage}[b]{1.0\linewidth}
  \centering
  \centerline{\includegraphics[width=8.3cm]{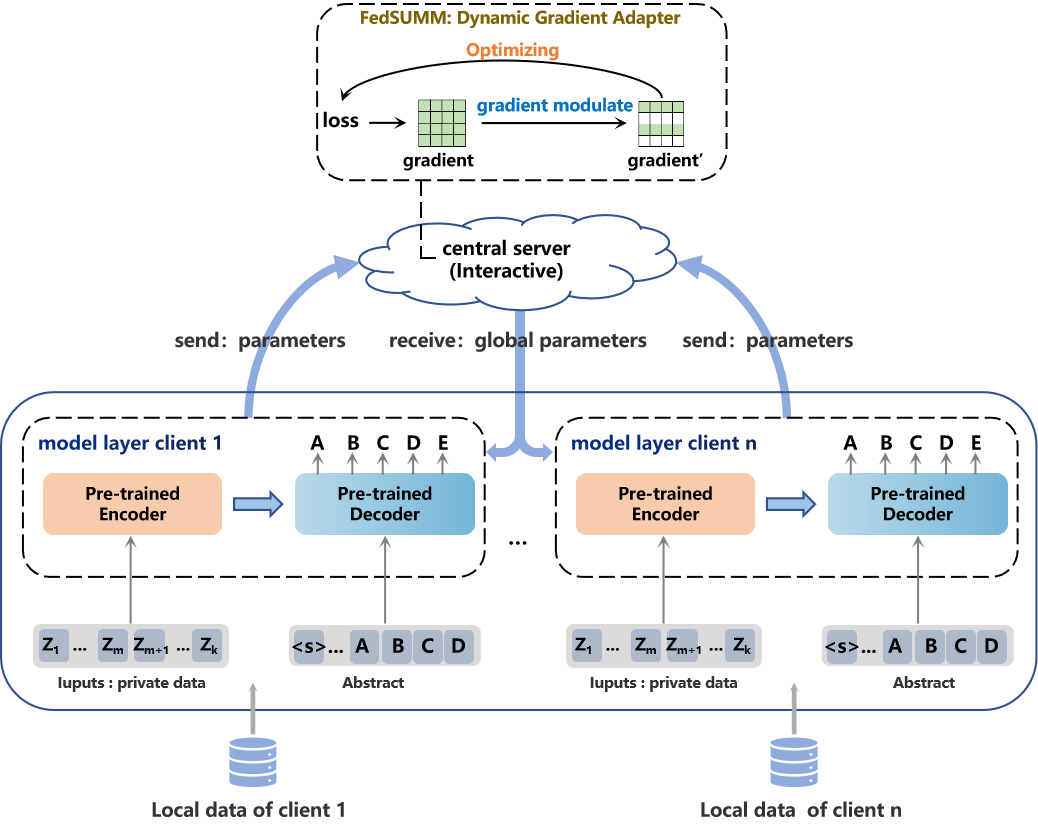}}
\end{minipage}
\caption{Federated Bart model structure, the federated aggregation algorithm uses our FedSUMM or replaces it with FedAvg.}
\label{fig:res}
\end{figure}

The FedAvg algorithm integrates decentralized models into a globally optimal model. The optimization objective (1) is Empirical Risk Minimization. $n$ is the sample size. $f^{}_{i}(w)$ represents the loss function of the model on the $i-th$ sample. 

\begin{equation}
    \min_{w\in \mathbb{R} ^{d}} \left [ F(w)=\frac{1}{n}\sum_{i=1}^{n}{f_{i}(w) }   \right ] 
\end{equation}

Assuming that $P^{}_{k}$ is a set of all samples of the $k-th$ model, let $n^{}_{k}$ = $\arrowvert P^{}_{k} \arrowvert $, then the objective function can be transformed into formula (2). In each iteration, the local model will perform a parameter update. In a batch, the formula (3) for the iterative update of the $k-th$ local model.

\begin{equation}
    F_{k} (w)=\frac{1}{n_{k} } \sum_{i\in P_{k} }^{} f_{i}  (w)
\end{equation}

\begin{equation}
    w_{k} \gets  w_{k-1} - \frac{\eta }{batch} \sum_{i\in batch}^{} \bigtriangledown f_{i} (w)
\end{equation}

\subsection{Federated CopyTransformer}
At the model layer, we also replace Bart with the classic Transformer to compare the performance of the pre-trained model and Transformer combined with federated learning. Junfan et al. [28] proposed a copy distribution that constrains the model to generating summaries. The attention distribution is modified by choosing probabilities that filter possible phrases as components of summaries. As shown in Fig.2, we use Transformer with a copy distribution attention mechanism: CopyTransformer (abbreviated as CT).

\begin{figure}[htb]
\begin{minipage}[b]{1.0\linewidth}
  \centering
  \centerline{\includegraphics[width=6cm]{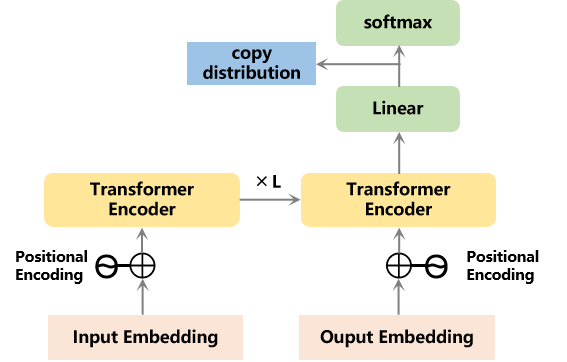}}
\end{minipage}
\caption{Model layer replaced by CopyTransformer.}
\label{fig:res}
\end{figure}

\subsection{FedSUMM}

To deal with the challenge of statistical diversity of text for personalized federated learning, especially when different local data have different semantic information. Our FedSUMM algorithm proposes maintaining different gradient information on the server to instruct clients to perform personalized model training. Moreover, for different clients using the weighted average method of the FedAvg algorithm limits the performance of clients with different data characteristics owing to the permutation invariance of neural network parameters\cite{tang2021sensory}\cite{chen2020parallel}. It shows the imbalance of the data distribution problem where the global model optimization only depends on the order of parameters \cite{zhang2022federated}, so we propose FedSUMM in Algorithm1 to address the problem of client drift. In the process of algorithm iteration, the dynamic gradient adapter will dynamically transfer the more appropriate gradient information back to the client to balance the generalization and personalization better. In each round of communication t, the central server receives the gradient parameter $G\in\{g^{}_{1},...,g^{}_{n}\}$ sent by the client, and the local training loss $L \in \{l^{}_{1},...,l^{}_{n}\}$. The central server updates the historical record memory gradients (MG) to ensure that the client's latest personalized customization gradient parameters use for each communication and sorts the MG in ascending order according to the loss of each client. 

\begin{algorithm}[!h]
\renewcommand{\algorithmicensure}{\textbf{clientUpdate(c , W):}}
\renewcommand{\algorithmicensure}{\textbf{gradient\_adapter(MG , Loss):}}
\caption{FedSUMM}
\label{alg:A}
\begin{algorithmic}
\renewcommand{\algorithmicrequire}{\textbf{Input:}}
\REQUIRE the communication round T is indexed by t; clients number C are indexed by c; total train step TS; B is batch size; the model is M; the learning rate is $\delta$.
\renewcommand{\algorithmicensure}{\textbf{gradient\_adapter(MG , Loss):}}
\ENSURE
\STATE \textbf{for} i and j = i + 1 in range(len(C)) \textbf{ do}
\STATE \quad sorts MG elements , where $l^{i}_{t+1} \leq l^{j}_{t+1} $ based on Loss
\STATE $g^{v}_{t+1} = MG[v] \ast \varepsilon \ast \rho$ , when $\left \| \bigtriangledown _{w} f_{i}(w)  \right \| \ne 1$
\renewcommand{\algorithmicensure}{\textbf{Server executes:}}
\ENSURE
\STATE initialize $w^{}_{0}$ and memory gradient $MG = []$
\FOR {each round $t=1,2 \dots$ }
\STATE $S^{}_{t} \leftarrow $ ( random allocation of C clients )
\STATE sends model $w^{}_{t}$ to clients side
\STATE \textbf{for} each client $c \in S^{}_{t}$ \textbf{in parallel do}
\STATE \quad \leftline{$g^{c}_{t+1},l^{c}_{t+1},w^{c}_{t+1} \leftarrow ClientUpdate(c,w,g,l)$}\\
\STATE \textbf{end for}
\STATE $MG\_new \leftarrow gradient\_adapter(MG,Loss)$\\
\STATE $w^{c}_{t+1} \leftarrow w^{c}_{t+1} , MG\_new[c]$
\STATE Global aggregation: $w^{}_{t+1} \leftarrow \sum^{C}_{c = 1}\frac{1}{C}w^{c}_{t+1} $
\ENDFOR
\renewcommand{\algorithmicensure}{\textbf{ClientUpdate(c , w , g , l): } // local run on client c}
\ENSURE
\FOR {each local train step ts from 1 to TS}
\STATE \textbf{for} batch $ b \in B $ \textbf{do}
\STATE \quad Update personalized model: w $\leftarrow$ w - $\delta$ $\bigtriangledown$ M(w , b)
\ENDFOR
\end{algorithmic}
\end{algorithm}

\begin{table*}[!htbp]
\caption{\centering\scshape The Comparisons between the Proposed FedSUMM Algorithm and FedAvg. Blue Bold Indicates the Methods with the best Performances. the number of Participating Clients is 50. \# column: single and All Represent the Training Result of A Local Single Client or All Client Data Collection, Respectively. Fl Represents Traditional Federated Learning Methods, Ours Represents Our Proposed Personalized Federated Learning Methods. CT abbreviation for CopyTransformer.}
\resizebox{\textwidth}{28mm}{
\begin{tabular}{@{}cclllllllllllllll@{}}
\toprule
\multicolumn{1}{l}{} &  & \multicolumn{3}{c}{CSL} & \multicolumn{3}{c}{NLPCC} & \multicolumn{3}{c}{LCSTS} & \multicolumn{3}{c}{THUCNEWS} & \multicolumn{3}{c}{EDUCATION} \\
model & \# & \multicolumn{3}{c}{(R1 / R2 / RL)} & \multicolumn{3}{c}{(R1 / R2 / RL)} & \multicolumn{3}{c}{(R1 / R2 / RL)} & \multicolumn{3}{c}{(R1 / R2 / RL)} & \multicolumn{3}{c}{(R1 / R2 / RL)} \\ \midrule
\multirow{2}{*}{Bart} & single & \multicolumn{3}{c}{45.2 / 25.7 / 39.5} & \multicolumn{3}{c}{46.8 / 32.4 / 41.3} & \multicolumn{3}{c}{35.7 / 24.0 / 31.9} & \multicolumn{3}{c}{32.1 / 20.8 / 29.4} & \multicolumn{3}{c}{45.9 / 34.6 / 43.1} \\ 
 & all & \multicolumn{3}{c}{48.7 / 32.3 / 43.4} & \multicolumn{3}{c}{51.4 / 39.2 / 47.0} & \multicolumn{3}{c}{39.4 / 30.2 / 38.9} & \multicolumn{3}{c}{35.4 / 28.5 / 34.6} & \multicolumn{3}{c}{50.7 / 41.2 / 49.6} \\ \midrule
Bart+FedAvg & fl & \multicolumn{3}{c}{46.1 / 25.9 / 40.3} & \multicolumn{3}{c}{47.5 / 32.9 / 41.5} & \multicolumn{3}{c}{34.9 / 24.3 / 32.0} & \multicolumn{3}{c}{32.4 / 17.1 / 21.9} & \multicolumn{3}{c}{46.0 / 33.7 / 43.5} \\ \midrule
Bart+FedSUMM & ours & \multicolumn{3}{c}{\color{blue}\textbf{48.0} / \color{blue}\textbf{31.5} / \color{blue}\textbf{42.7}} & \multicolumn{3}{c}{\color{blue}\textbf{49.6} / \color{blue}\textbf{37.3} / \color{blue}\textbf{45.9}} & \multicolumn{3}{c}{\color{blue}\textbf{37.7} / \color{blue}\textbf{28.9} / \color{blue}\textbf{36.5}} & \multicolumn{3}{c}{\color{blue}\textbf{33.8} / \color{blue}\textbf{25.9} / \color{blue}\textbf{32.6}} & \multicolumn{3}{c}{\color{blue}\textbf{49.5} / \color{blue}\textbf{38.7} / \color{blue}\textbf{46.8}} \\ \midrule
\multirow{2}{*}{CT} & single & \multicolumn{3}{c}{40.7 / 20.5 / 33.6} & \multicolumn{3}{c}{43.9 / 28.2 / 38.7} & \multicolumn{3}{c}{29.9 / 18.4 / 27.6} & \multicolumn{3}{c}{25.2 / 12.4 / 21.4} & \multicolumn{3}{c}{31.7 / 21.2 / 27.9} \\ 
 & all & \multicolumn{3}{c}{44.8 / 23.4 / 37.3} & \multicolumn{3}{c}{45.8 / 29.3 / 40.3} & \multicolumn{3}{c}{34.6 / 21.5 / 31.8} & \multicolumn{3}{c}{30.3 / 16.4 / 26.2} & \multicolumn{3}{c}{43.9 / 32.3 / 40.3} \\ \midrule
CT+FedAvg & fl & \multicolumn{3}{c}{41.9 / 21.8 / 34.1} & \multicolumn{3}{c}{44.5 / 28.5 / 39.2} & \multicolumn{3}{c}{31.4 / 19.1 / 28.6} & \multicolumn{3}{c}{25.4 / 12.8 / 22.2} & \multicolumn{3}{c}{37.9 / 26.9 / 34.8} \\ \midrule
CT+FedSUMM & ours & \multicolumn{3}{c}{\color{blue}\textbf{43.1} / \color{blue}\textbf{22.7} / \color{blue}\textbf{35.8}} & \multicolumn{3}{c}{\color{blue}\textbf{45.2} / \color{blue}\textbf{28.8} / \color{blue}\textbf{39.9}} & \multicolumn{3}{c}{\color{blue}\textbf{33.8} / \color{blue}\textbf{20.7} / \color{blue}\textbf{30.4}} & \multicolumn{3}{c}{\color{blue}\textbf{27.2} / \color{blue}\textbf{15.3} / \color{blue}\textbf{24.7}} & \multicolumn{3}{c}{\color{blue}\textbf{39.7} / \color{blue}\textbf{28.7} / \color{blue}\textbf{36.3}} \\ \midrule

\end{tabular}
}
\end{table*}

In the FedSUMM algorithm, the optimization objective is still empirical risk minimization. Suppose the loss function of each participant is $L(\theta)$. We can define the parameter $\theta$ as an updated trajectory that varies with the number of communication rounds t $\in$ (0, T). Then the update change rate can be expressed as the formula (4).

\begin{equation}
    \frac{d}{d_{t} } L(\theta (t))=\left \{ \bigtriangledown _{\theta } L(\theta (t)) , \dot{\theta } (t) \right \} 
\end{equation}

We propose to adaptively modulate the local model's gradient during global optimization by monitoring the discrepancy of each local model to the learning text heterogeneous semantics. Here we design the discrepancy ratio $\rho$ as formula (5), which dynamically monitors the discrepancy between clients and the global model due to text heterogeneity. In formula (6), $\sigma _{c,t}$ and $\sigma _{g,t}$ as the approximated predictions of the local model and the global model, respectively. It estimates the performance after normalizing raw data and gradient parameters after global optimization.

\begin{equation}
    \rho  =\frac{ {\textstyle \sum_{t\in T}^{} \sigma _{c,t} }^{} }{{\textstyle \sum_{t\in T}^{} \sigma _{g,t} }^{}}
\end{equation}

\begin{equation}
    \sigma _{c,t} =\frac{exp(w_{i}^{c} )}{ {\textstyle \sum_{j}^{}}exp(w_{j}^{c}) } ; \sigma _{g,t} =\frac{exp(w_{i}^{g} )}{ {\textstyle \sum_{j}^{}}exp(w_{j}^{g}) }
\end{equation}

A formula (7), the $\theta$, is a trainable parameter for model $f(w)$. Moreover, the training loss for a single sample is expressed as $loss(f_{i}(w),s)$. For each client model $f(w)$ we compute a item of gradient adapter $\left \| \bigtriangledown _{w} f_{i}(w)  \right \|$. When $\left \| \bigtriangledown _{w} f_{i}(w)  \right \| \ne 1$, the adapt is made, where $\varepsilon$ is a hyper-parameter to control the degree of modulation for gradient adapter. Furthermore, the farther the distance is from 1, the greater the adept.

\begin{equation}
\begin{aligned}
 arg\min_{\theta } L(\theta ) &= \mathbb{E}_{(w,s)\sim D}[loss(f_{i}(w),s)] \\
 &+ \varepsilon \cdot \rho \cdot \frac{d}{d_{t} } L(\theta (t)) \cdot \max ( \left \|\bigtriangledown _{w} f_{i}(w)  \right \| ,1) 
\end{aligned}
\end{equation}

Intuitively, only federated learning can protect privacy, but gradient information will also data breach \cite{Qiuksem}\cite{Liuicassp}, the transmission of gradients can also cause leakage of data distribution. In order to prevent gradient leakage, we adopt differential privacy to add noise to gradient information. 

Differential privacy is a method that allows data to be used for any analysis and ensures that users are not adversely affected \cite{zhao2022fldp}. Differential privacy is defined as inequality (8). Suppose there is a random algorithm $M$, $S$ is the set of all possible output results for $M$, and $\mathbb{P} _{r}$ represents the probability. For any two adjacent data sets, if inequality (8) is satisfied, then the algorithm $M$ is considered to have $\epsilon -$ differential privacy protection, where $\epsilon$ is the differential privacy budget, which is used to ensure the probability that the output of the random algorithm is consistent when a record is added or subtracted from the data set.

\begin{equation}
    \mathbb{P} _{r} \left [ M(D)\in S \right ] \le e ^{\epsilon } \left [ M(D^{\prime } ) \in S \right ] +  \delta 
\end{equation}

The strength of the added noise is related to the median $M$ (9) of the user gradient update norm. The central server adds Gaussian noise N and then does a global optimization (10). $K$ is the number of clients. $\left \{ \sigma  \right \}_{t=0}^{T}$ is the set of variances for the Guassian mechanism. $\zeta$ is the norm of update for each client.

\begin{equation}
    M = \left \{ \zeta ^{k}  \right \}_{k\in S_{t} }
\end{equation}

\begin{equation}
    W_{t+1} \longleftarrow W_{t} + \nonumber \frac{1}{m} \left ( \sum_{k=1}^{K}\Delta W_{t+1}^{k} \/max(1,\frac{\zeta ^{k} }{M})+\mathcal{N}(0,M^{2}·\sigma ^{2} ) \right)
    \eqno(10) 
\end{equation}

\section{Experiments}
\subsection{Experimental Settings}\label{AA}

\noindent \textbf{Datasets: }CSL is Chinese scientific literature dataset. The NLPCC dataset is widely used in the public Chinese text summarization evaluation, which contains about 50,000 real-long news articles paired. LCSTS is a large-scale Chinese short text summarization dataset constructed from the microblogging website Sina Weibo. THUCNEWS is 0.74 million news documents. EDUCATION is a dataset about mainstream vertical media in the education and training industry, which consists of over 2 million real Chinese short texts.

\noindent \textbf{Evaluation Metrics: }We evaluate models using standard full-length ROUGE F1 \cite{lin2004rouge}, which is currently the most common evaluation index for text summarization. ROUGE-1 (R1), ROUGE-2 (R2), and ROUGE-L (RL) refer to the matches of unigrams, bigrams, and the longest common subsequence algorithm, respectively. The ROUGE-N evaluation metrics calculates the n-gram recall rate and precision rate of the generated summary and the corresponding reference summary. The ROUGE-L evaluation metrics calculates the F-measure based on the longest common sequence of two text units.

\subsection{Experimental Results}

We conducted different experiments using five datasets, and the overall experimental performance is shown in Table I. We applied CT (abbreviation for CopyTransformer) and a pre-trained model Bart as base models for the text summarization. We use Bart-Base-Chinese pre-trained model open source by Fudan University\cite{shao2021cpt}. On this basis, the classical FL algorithm FedAvg and our proposed PFL algorithm FedSUMM are applied for distributed training in the model column of Table I, represented as Bart+FedAvg,  CT+FedAvg, Bart+FedSUMM, and CT+FedSUMM, respectively.

\begin{figure}
  \begin{subfigure}[t]{0.5\linewidth}
    \centering
    \includegraphics[scale=0.3]{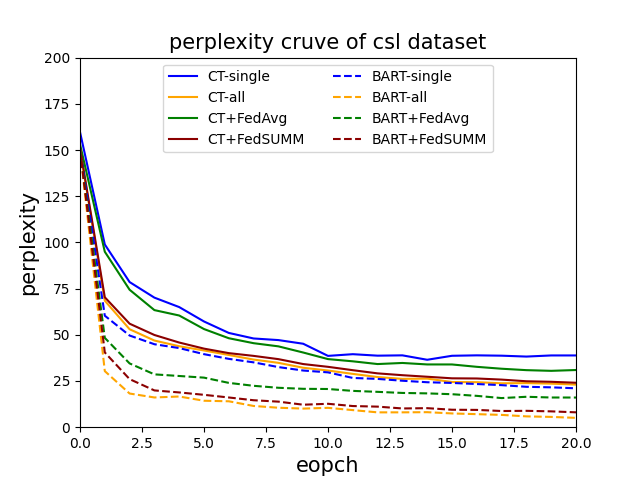}
    \caption{Perplexity on CSL}
    \label{fig:side:a}
  \end{subfigure}%
  \begin{subfigure}[t]{0.5\linewidth}
    \centering
    \includegraphics[scale=0.3]{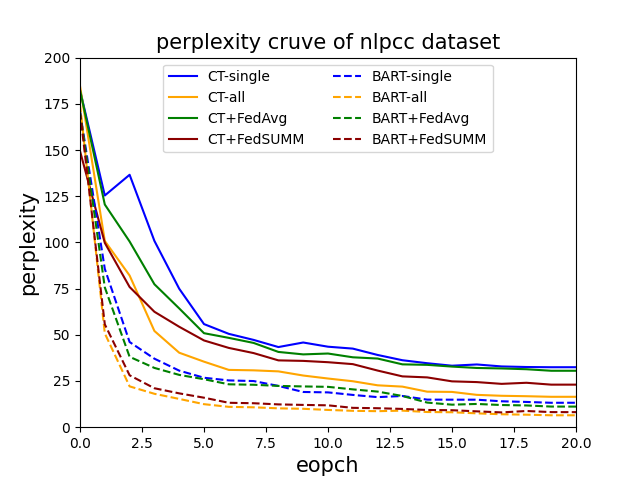}
    \caption{Perplexity on NLPCC}
    \label{fig:side:a}
  \end{subfigure}%
  \qquad
  \begin{subfigure}[t]{0.5\linewidth}
    \centering
    \includegraphics[scale=0.3]{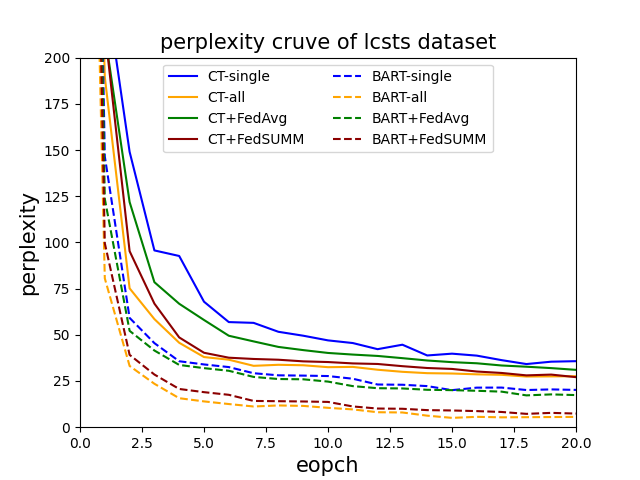}
    \caption{Perplexity on LCSTS}
    \label{fig:side:a}
  \end{subfigure}%
  \begin{subfigure}[t]{0.5\linewidth}
    \centering
    \includegraphics[scale=0.3]{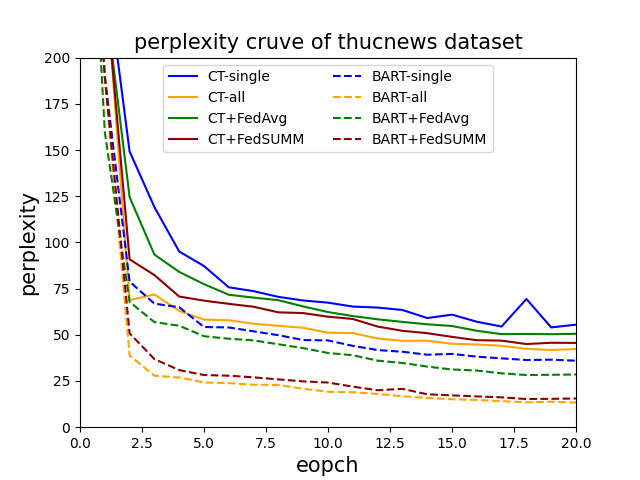}
    \caption{Perplexity on THUCNEWS}
    \label{fig:side:a}
  \end{subfigure}%
  \qquad
  \begin{subfigure}[t]{1\linewidth}
    \centering
    \includegraphics[scale=0.3]{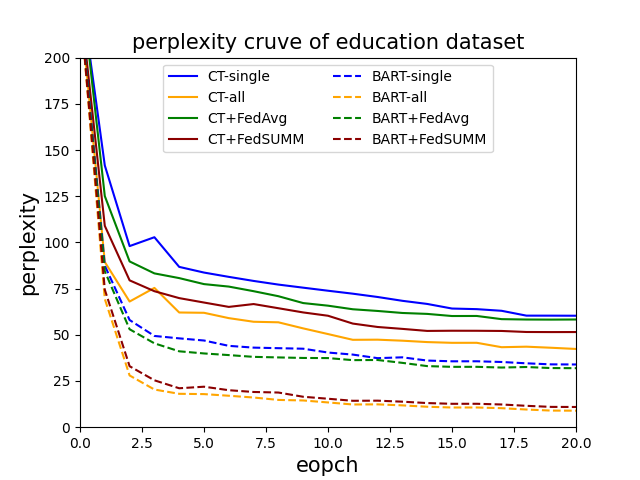}
    \caption{Perplexity on EDUCATION}
    \label{fig:side:a}
  \end{subfigure}%
\caption{Perplexity estimation for different methods on five datasets.}
\end{figure}

\begin{figure}
  \begin{subfigure}[t]{0.5\linewidth}
    \centering
    \includegraphics[scale=0.3]{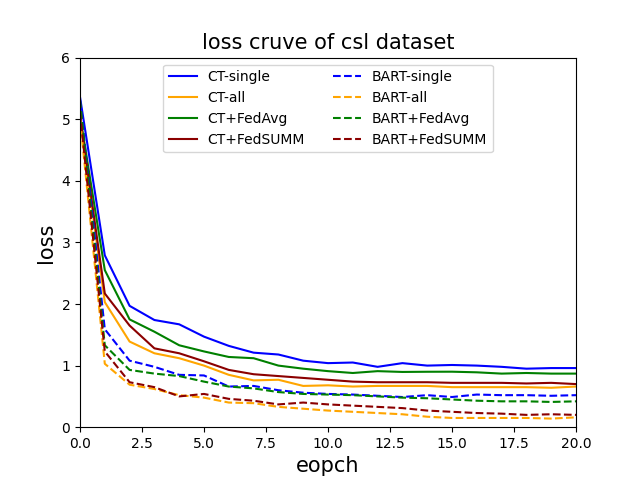}
    \caption{Loss on CSL}
    \label{fig:side:b}
  \end{subfigure}%
  \begin{subfigure}[t]{0.5\linewidth}
    \centering
    \includegraphics[scale=0.3]{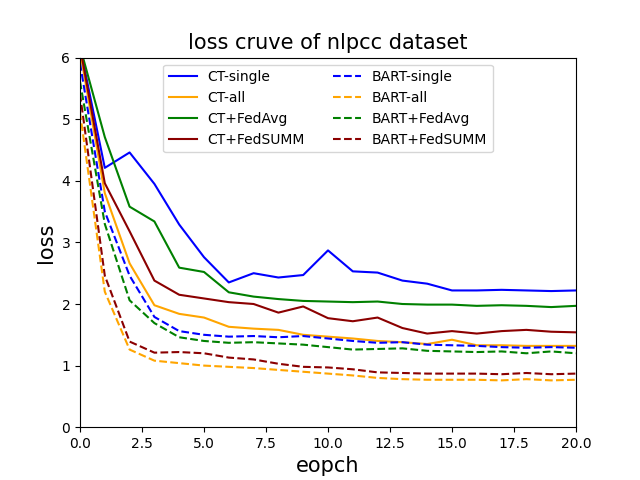}
    \caption{Loss on NLPCC}
    \label{fig:side:b}
  \end{subfigure}%
  \qquad
  \begin{subfigure}[t]{0.5\linewidth}
    \centering
    \includegraphics[scale=0.3]{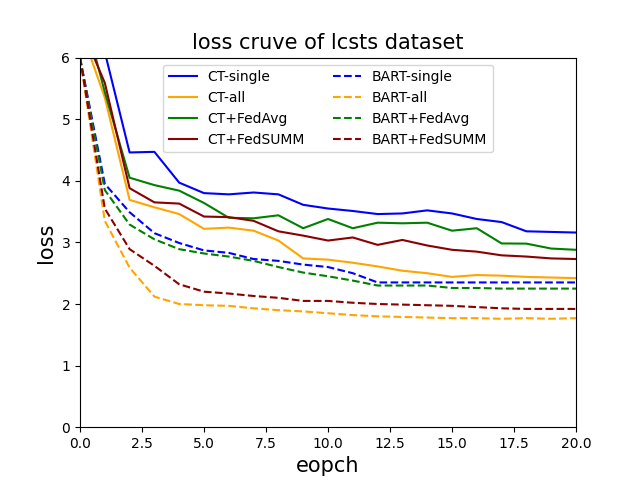}
    \caption{Loss on LCSTS}
    \label{fig:side:b}
  \end{subfigure}%
  \begin{subfigure}[t]{0.5\linewidth}
    \centering
    \includegraphics[scale=0.3]{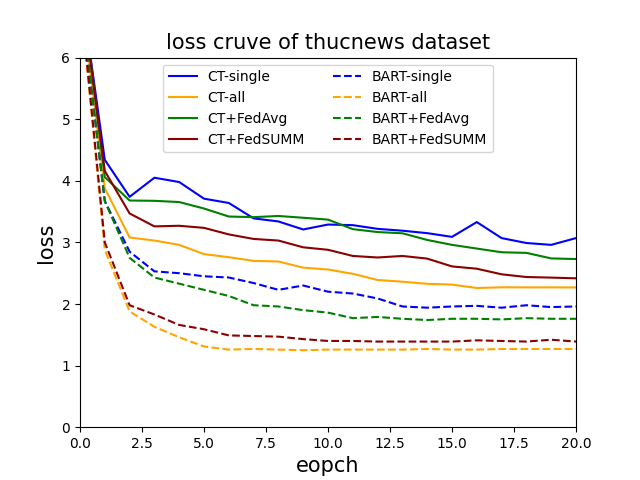}
    \caption{Loss on THUCNEWS}
    \label{fig:side:b}
  \end{subfigure}%
  \qquad
  \begin{subfigure}[t]{1\linewidth}
    \centering
    \includegraphics[scale=0.3]{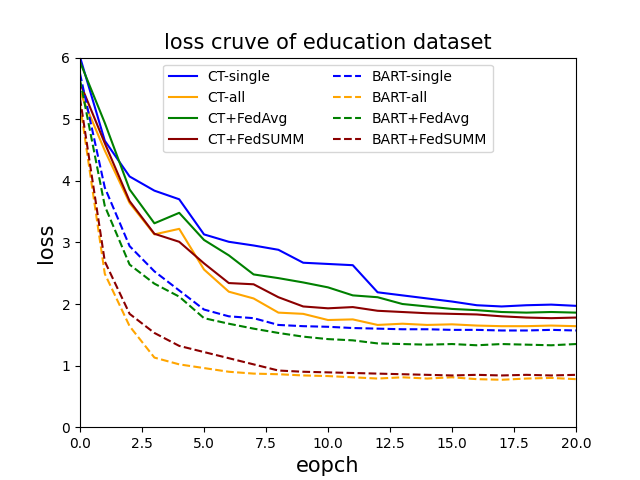}
    \caption{Loss on EDUCATION}
    \label{fig:side:b}
  \end{subfigure}%
\caption{Loss changes during training for different methods on five datasets.}
\end{figure}

We can observe that our methods FedSUMM outperform the FedAvg. As in Table I, all of the \# columns indicate the performance on rogue FedSUMM closer to the effect of collecting all client local data in generalization performance. In order to obtain more suitable global optimization semantics for the local model, the gradient information is dynamically adjusted by a gradient adaptor, and the global model is instructed to pass more appropriate generalization loss and gradient parameters to the local client. Overall, the Application of our algorithm FedSUMM on CT achieves about 1.2\%, 0.9\%, and 1.7\% improvements across ROUGE score variants on CSL data compared to FedAvg. Compared with FedAvg, the combination of FedSUMM and pre-trained model Bart achieves about 1.9\%, 5.6\%, and 2.4\% improvement in the ROUGE scores R1, R2, and RL of the CSL data, respectively. As shown in the single row in column \# of Table I, compared with FedSUMM. This personalized federated learning algorithm solves text heterogeneity. Due to the independent form of physical data islands, the performance obtained by a single training of the local model has a significant gap compared to FedSUMM.

As shown in Fig.3, our experiment compares the perplexity of the eight models during training and measures the quality indicators of the eight models for five datasets. CT-single and BART-single represent a single training of the CopyTransformer and Bart model training using local data, respectively. CT-all and BART-all indicate that the CopyTransformer and Bart model training uses the data integrated by all parties. It mainly estimates the probability of a sentence based on each word and normalizes the sentence length. The smaller the perplexity, the higher the probability of each word in the abstract, Indicating that the generated summary sentences are of better quality. We can observe that FedSUMM has faster fitting speed, lower perplexity, and better smoothness of the generated summaries.

Besides, we can also see the loss trend of the training process in Fig.4. we can observe that the convergence rate of the FedAvg is lower than that of our personalized federated learning FedSUMM. In each iteration of model training, our methods fully consider the semantic deviation caused by text statistical heterogeneity. Dynamically modulate the gradient to seek a balance between personalized local client optimization and global model generalization to fit the text summarization task faster.

\begin{figure}[h]
 \begin{subfigure}[t]{0.5\linewidth}
    \centering
    \includegraphics[scale=0.3]{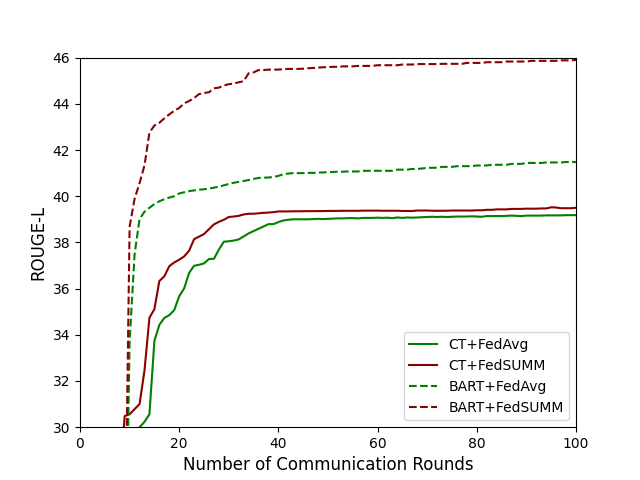}
    \caption{communication}
    \label{fig:side:a}
  \end{subfigure}%
  \begin{subfigure}[t]{0.5\linewidth}
    \centering
    \includegraphics[scale=0.3]{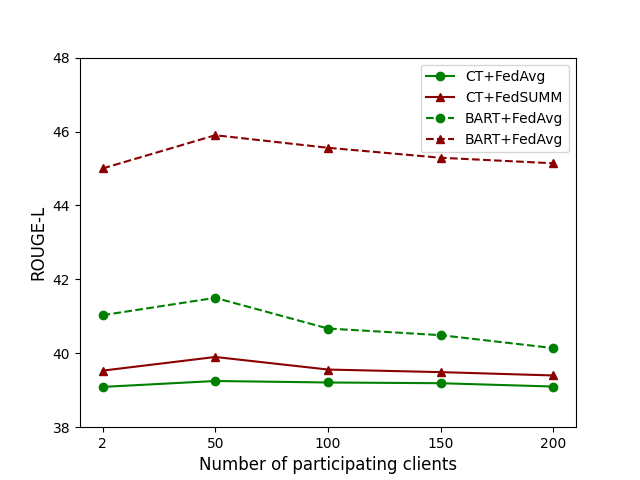}
    \caption{clients number}
    \label{fig:side:a}
  \end{subfigure}%
  \caption{ROUGE-L F1 scores of different models under different settings on the number of communication rounds (a) and the number of clients (b) in one iteration.}
	\label{da_chutian}
\end{figure}

As shown in Fig.5 (a),  from which we can observe that our FedSUMM algorithm can converge better with the increase in communication rounds. Compared with traditional federated learning algorithms FedAvg, such as Bart+FedAvg and CT+FedAvg, our FedSUMM algorithm can achieve better performance in fewer communication rounds. Besides, it can reduce communication time, as shown in the curves of CT+Fedsumm and Bart+FedSUMM in Fig.5 (a). We observe in Fig.5 (b) that increasing the number of clients has a different effect on the performance of two federated learning methods, FedAvg and FedSUMM. There is a slight increase in ROUGE-L performance when the number of client participants increases between 2 to 200. Both methods show a downward trend as the number of clients increases (when more significant than 50) because Chinese text data is limited, and the more parties involved, the less data we have locally. Although partial participation reduces the convergence speed of FedAvg and FedSUMM, optimal solutions including local constraint terms can still be obtained. Increasing the number of client participants can achieve better performance at a particular data level.

\begin{figure}[h]
    \begin{subfigure}[t]{1\linewidth}
    \centering
    \includegraphics[scale=0.3]{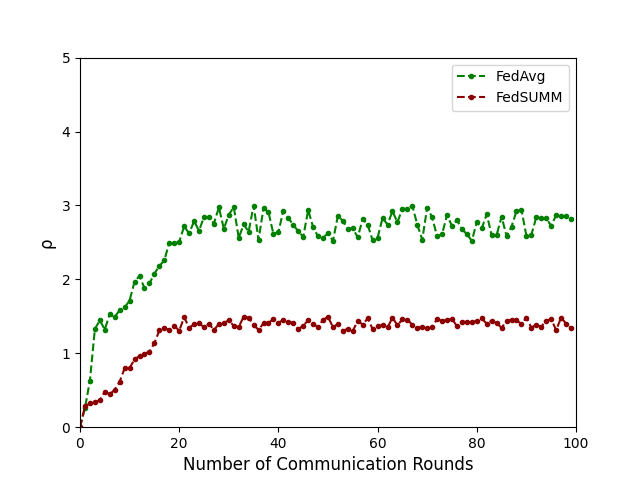}
    \end{subfigure}%
	\caption{Change of discrepancy ratio $\rho$ between FedAvg and FedSUMM during communication.}
	\label{da_chutian}
\end{figure}

More importantly, it can be noticed from Fig.6 that the discrepancy ratio $\rho$ obviously decreases after applying our proposed personalized federated learning with the dynamic gradient adapter algorithm FedSUMM, which is strong evidence for the effectiveness of our algorithm. From Fig.6, we can observe that the client drift caused by text heterogeneity will become more pronounced, when the number of communication rounds increases. Suitable through personalized federation settings, our FedSUMM algorithm fits this heterogeneous deviation and reduces the discrepancy ratio to a value near 1. From a statistics point of view, the client's drift probability is lower when the discrepancy ratio is closer to 1.

\begin{table}[!ht]
\centering
\renewcommand\arraystretch{1.3}
\caption{\scshape Experimental Results of Ablation Study.}
    \setlength{\tabcolsep}{3mm}{	
    \begin{tabular}{llll}
    \toprule
        \textbf{Model} & \textbf{CSL} & \textbf{THUNEWS} & \textbf{EDUCATION} \\ \hline
        Bart & 39.5 & \makecell[c]{29.4} & \makecell[c]{43.1} \\ 
        + CD & 39.7 & \makecell[c]{30.1} & \makecell[c]{43.6} \\ 
        + CD \& FedAvg & 41.2 & \makecell[c]{27.5} & \makecell[c]{44.6} \\ 
        + CD \& FedSUMM & \textbf{42.8} & \makecell[c]{\textbf{32.6}} & \makecell[c]{\textbf{47.0}} \\ \bottomrule
    \end{tabular}}
\end{table}

In addition, we conduct a comprehensive ablation study to demonstrate FedSUMM's contributions. Intuitively, the discrepancy brought about by text heterogeneity will lead to the semantic drift of the encoding process. Then, this kind of semantic drift will play a very influential role in reorganizing the attention probability distribution through the copy distribution attention mechanism (abbreviation as CD) at the decoder. The smaller the degree of improvement, the stronger the capability of federated learning algorithm to reduce the discrepancy ratio of text heterogeneity.

The experimental results are shown in Table II. ROUGE-L F1 scores of different ablation studies. From these results, we can observe that the FedSUMM algorithm has no significant improvement in performance after adding CD, indicating that in the process of federated learning, the FedSUMM algorithm can deal well with the semantic drift caused by text heterogeneity, so that in the process of decoding, the demand for probability recombination of attention distribution is not obvious. As shown in Table I, the RL F1 scores of the Bart+FedAvg method on the three datasets, CSL, THUCNEWS, and EDUCATION, are 40.3, 21.9, and 43.5, respectively. A boost of 0.9, 5.6, and 1.1 when adding CD, respectively. Therefore, the increase of the discrepancy ratio increases the possibility of semantic restructuring in the decoding, so the role of CD is also reflected.

\section{Conclusion}
This paper proposes a method of distributed federated learning with Gaussian differential privacy to address the data privacy problem of scarce text summarization data. Experimental results show that the federated models outperform existing state-of-the-art methods, which only utilize local data. In addition, the results suggest that a dynamic gradient adapter should be used to address the problem of text heterogeneity for PFL. From the analysis of the experimental results, the FedSUMM method can better balance the generalization of the global model and personalization of the local client, while solving the global semantic deviation of different client texts during model optimization.

\section*{Acknowledgment}
Supported by the Key Research and Development Program of Guangdong Province (grant No. 2021B0101400003) and Corresponding author is Jianzong Wang (jzwang@188.com).

\bibliographystyle{unsrt}
\bibliography{reference.bib}

\vspace{12pt}

\end{document}